\ifdefined\pdfminorversion
\fi
\documentclass[10pt,twocolumn,letterpaper]{article}
\usepackage[OT1]{fontenc}
\usepackage[pagenumbers]{cvpr}
\usepackage{multirow}
\usepackage{colortbl}
\usepackage{tabularx}
\DeclareMathSizes{7.25}{7.25}{5}{5} 
\definecolor{ablationgray}{gray}{0.95}
\definecolor{cvprblue}{rgb}{0.21,0.49,0.74}
\usepackage[pagebackref,breaklinks,colorlinks,allcolors=cvprblue,pdfusetitle]{hyperref}

\title{AgroVG+: Generalized Visual Grounding in Agriculture with Region-Supervised Action Generation}
\author{%
    Haocheng Li\textsuperscript{1}\quad
    Juepeng Zheng\textsuperscript{2}\quad
    Guilong Xiao\textsuperscript{1}\quad
    Kaiqi Du\textsuperscript{1}\\
    Gengmeng Pu\textsuperscript{1}\quad
    Haohuan Fu\textsuperscript{3}\quad
    Jianxi Huang\textsuperscript{4}\\[4pt]
    \normalsize \textsuperscript{1}\,China Agricultural University\quad
    \normalsize \textsuperscript{2}\,Sun Yat-sen University\\
    \normalsize \textsuperscript{3}\,Tsinghua University\quad
    \normalsize \textsuperscript{4}\,Southwest Jiaotong University
}
\hypersetup{pdfauthor={Haocheng Li, Juepeng Zheng, Guilong Xiao, Kaiqi Du,
    Gengmeng Pu, Haohuan Fu, Jianxi Huang}}

\begin{document}
\maketitle
\suppressfloats[t]
\begin{figure}[t]
    \centering
    \includegraphics[width=\linewidth]{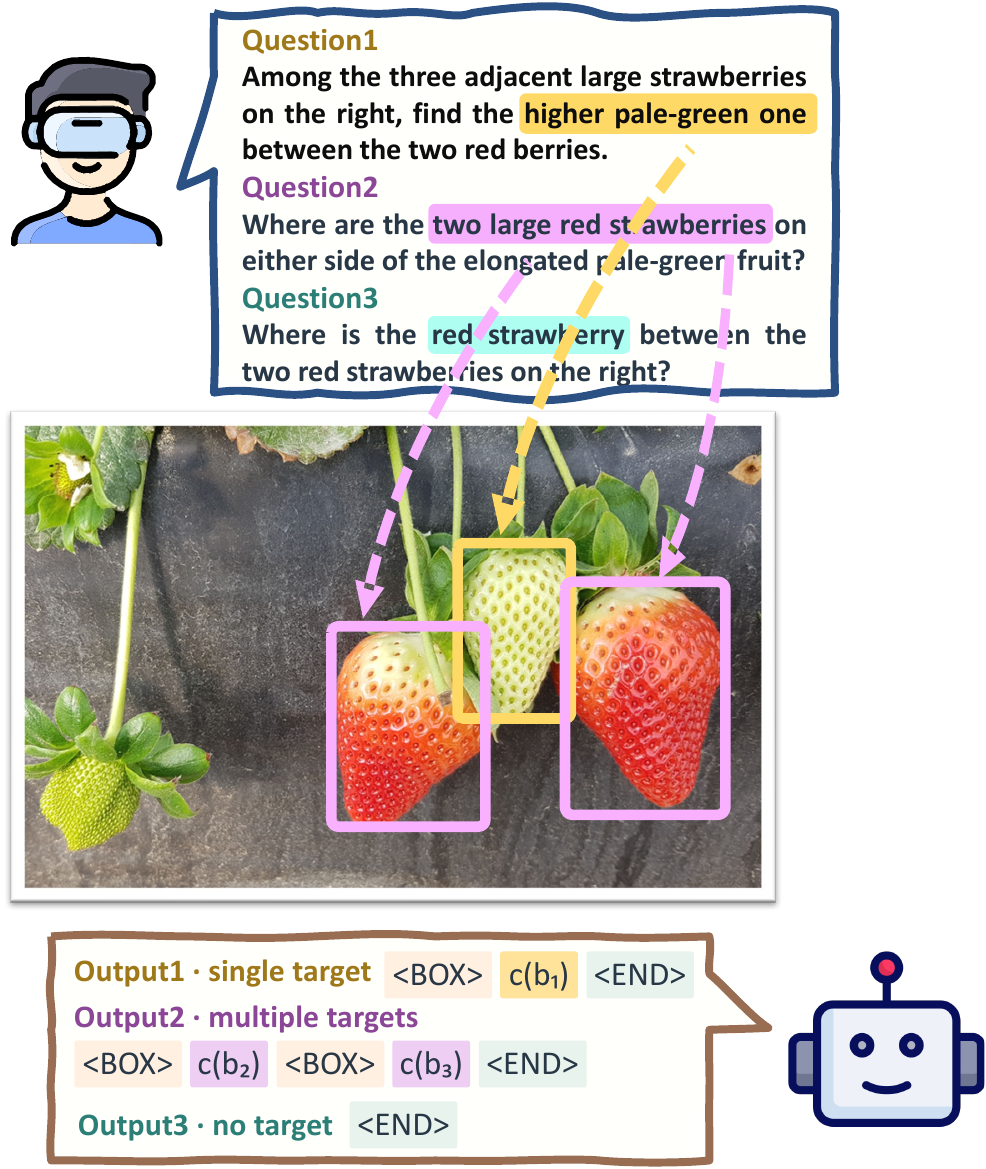}
    \caption{\textbf{One image, three queries, different target sets.}
    The queries refer to a single pale-green strawberry (yellow), two flanking red strawberries (purple), or no matching red strawberry between them.
    ReSAG represents each target set as \texttt{<BOX>} actions and coordinate strings $c(b_i)$, followed by \texttt{<END>}; an initial \texttt{<END>} denotes the empty set.}
    \label{fig:teaser}
\end{figure}

\begin{abstract}
Generalized visual grounding in agriculture requires localizing all objects that match a natural-language description and recognizing when no matching target is present.
We introduce \textbf{AgroVG+}, a dataset whose referring expressions are all written and reviewed by agricultural experts based on image content and corresponding target sets.
The dataset comprises 7,404 images and 10,081 query records across six agricultural scene families, covering single-target, multi-target, and target-absent cases.
We also propose \textbf{ReSAG} (Region-Supervised Action Generation), a framework that couples regional evidence supervision with explicit box-generation and termination decisions.
ReSAG assigns each target an equally weighted supervision term on a shared spatial evidence distribution computed before answer generation.
The BOX/END protocol exposes a choice between generating another box and terminating the response at box boundaries, while coordinates are generated using the original language vocabulary.
Regional and sequence objectives jointly train a shared language decoder.
On AgroVG+, ReSAG outperforms coordinate-only fine-tuning on single- and multi-target exact-set accuracy, and achieves higher aggregate accuracy than either component alone in ablations.
\end{abstract}
\section{Introduction}
\label{sec:intro}
Language-guided perception in agriculture requires a model to localize every object that matches a referring expression and recognize when no matching object is present (Fig.~\ref{fig:teaser}). A request to locate weeds, for example, may refer to a single plant, several plants scattered among crops, or none in the image. Repetitive visual patterns, partial occlusion, and visually similar vegetation make it difficult to distinguish matching instances from their surroundings~\cite{liAgroVGLargeScaleMultiSource2026,haghighatMultilabelInstancelevelGeneralised,wangDepthAwareConcealedCrop2024}. Generalized visual grounding captures this setting by treating localization as the prediction of a complete target set whose size depends on both the image and the expression~\cite{liu_gres_2023,he_grec_2023,dingGRExGeneralizedReferring2026a}.

Recent advances in multimodal large language models (MLLMs)~\cite{liBLIP2BootstrappingLanguageImage,daiInstructBLIPGeneralpurposeVisionLanguage,liuVisualInstructionTuning} have brought language understanding and spatial prediction into a shared generative framework~\cite{bai_qwen25vl_2025,bai_qwen3vl_2025,liuVISIONREASONERUNIFIEDREASONINGINTEGRATED2026}. By expressing object locations as coordinate sequences, these models perform visual grounding through the same autoregressive interface used for language generation~\cite{chenShikraUnleashingMultimodal2023}. Although this formulation supports variable-length outputs~\cite{pengGROUNDINGMULTIMODALLARGE2024a}, coordinate-only fine-tuning supervises answer tokens without directly constraining how spatial evidence is distributed across target regions. Complete-set prediction also requires deciding whether to emit a box or terminate the response, both at its beginning and after each completed box~\cite{chenPIX2SEQLANGUAGEMODELING2022}. These observations motivate coupling regional evidence supervision with explicit continuation and termination actions within an autoregressive grounding framework. 

Bounding-box annotations provide two complementary training signals for complete-set prediction. Localizing only prominent instances can omit less salient targets that also satisfy the expression, whereas generating additional boxes can introduce false positives. Each annotated box specifies the spatial region of a matching instance, while the complete target set provides supervision for continuing box generation and terminating once all annotated targets have been generated. An empty target set corresponds to termination before any box is generated.

Building on this observation, we introduce \textbf{ReSAG} (Region-Supervised Action Generation) for generalized visual grounding. ReSAG assigns each target an equally weighted loss term on a shared spatial evidence distribution computed from the image--expression prefix before answer generation. A target with low regional evidence therefore retains its own
penalty even when other targets receive strong evidence. Its BOX/END protocol restricts the decoder to \texttt{<BOX>} or \texttt{<END>} at the start of the answer and after each completed box, while coordinates are generated using the original language vocabulary. The regional and sequence losses jointly update the shared language decoder, allowing spatial supervision to shape the model that generates coordinates and predicts continuation and termination actions.

To support learning and evaluating generalized visual grounding in agriculture, we introduce \textbf{AgroVG+}, a dataset pairing natural-language expressions with complete target sets. Building on AgroVG~\cite{liAgroVGLargeScaleMultiSource2026}, we reselect images based on visual quality and expand the image collection through additional sampling from the original agricultural sources. A team of eight agricultural-domain experts writes all referring expressions from image content and manually reviews each expression and its corresponding target set. The resulting dataset contains 7,404 images and 10,081 bilingual query records across six agricultural scene families, with English and Chinese expressions sharing the same target sets. Its single-target, multi-target, and target-absent queries provide a unified setting for evaluating instance localization, complete-set recovery, and target-absence recognition.

Experiments on AgroVG+ show that ReSAG improves complete-set prediction for both single-target and multi-target queries compared with the evaluated baselines.
Ablation studies further show that combining regional evidence supervision with the BOX/END protocol yields higher aggregate accuracy than using either component alone.
These results support jointly learning spatial evidence and explicit box-generation and termination decisions for generalized agricultural grounding.

In summary, our contributions are as follows:
\begin{itemize}
    \item We introduce \textbf{AgroVG+}, an agricultural grounding dataset with expert-written referring expressions and manually reviewed target sets covering single-target, multi-target, and target-absent queries.
    
    \item We propose \textbf{ReSAG}, a framework that couples regional evidence supervision with explicit BOX/END actions for box generation and termination through joint training of a shared language decoder.
    
    \item We evaluate ReSAG against grounding baselines on AgroVG+ and demonstrate improved single- and multi-target exact-set accuracy. Controlled ablations and qualitative comparisons examine the effects of regional supervision and the BOX/END protocol.
\end{itemize}

\section{Related Work}
\label{sec:related_work}

\begin{figure*}[t]
    \centering
    \includegraphics[width=\textwidth]{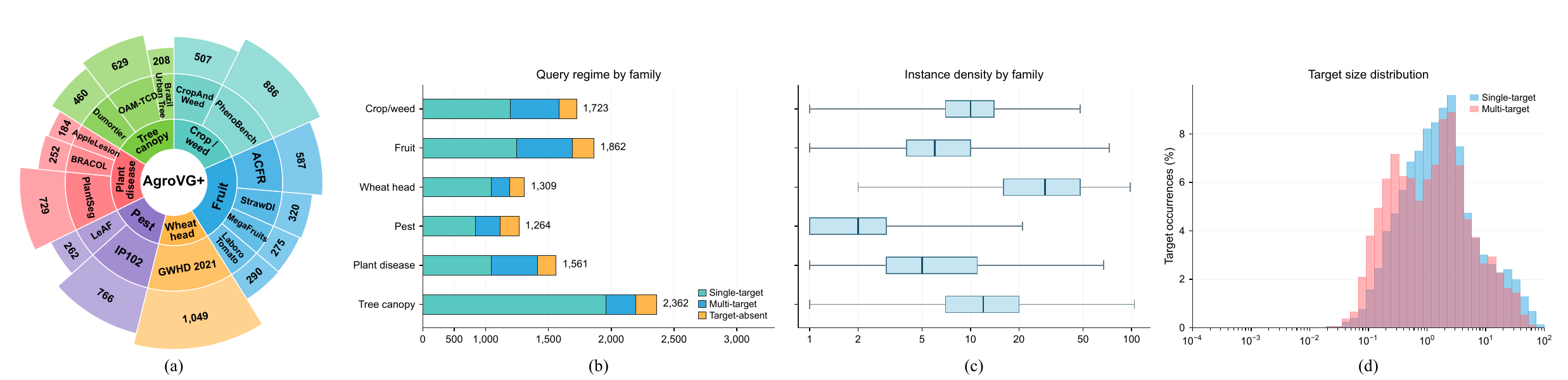}
    \caption{\textbf{Dataset statistics of AgroVG+.}
Panels summarize source/family composition, query-regime counts,
per-image instance-count distributions, and queried target-size distributions.
The six scene families include single-target, multi-target, and target-absent queries.}
    \label{fig:agrovg_plus_statistics}
\end{figure*}

\noindent\textbf{Agricultural Visual Grounding and Benchmarks.}
Existing agricultural datasets, including IP102~\cite{wu_ip102_2019}, Agriculture-Vision~\cite{chiuAgricultureVisionLargeAerial2020a}, and PhenoBench~\cite{weyler_phenobench_2024}, support fine-grained recognition and spatial analysis across diverse scenes~\cite{singhPlantDocDatasetVisual2020,hani_minneapple_2020,olsenDeepWeedsMulticlassWeed2019}.
For vision-language tasks, AgroGPT~\cite{awaisAgroGPTEfficientAgricultural2025} explores agricultural instruction tuning, while AgroBench~\cite{shinoda_agrobench_2025} and AgroMind~\cite{li_can_2025} evaluate multimodal capabilities in agriculture.
More directly related to localization, gRef-CW and its associated Weed-VG framework~\cite{haghighatMultilabelInstancelevelGeneralised} investigate crop and weed grounding, while AgroGround~\cite{alshehhiAgroGroundMultiGranularityGrounded2026} couples agricultural recognition with region localization.
AgroVG~\cite{liAgroVGLargeScaleMultiSource2026} evaluates single-target, multi-target, and target-absent grounding across agricultural sources using annotation-grounded query templates, including category, spatial, and size-based descriptions.
All referring expressions in AgroVG+ are written by agricultural experts based on the visible content of each scene, drawing on object appearance, spatial context, and relationships among instances.
Each expression and its complete target set are jointly reviewed for consistency with the image.

\noindent\textbf{Generalized Visual Grounding.}
Generalized grounding extends conventional single-instance referring expression comprehension~\cite{kazemzadehReferItGameReferringObjects2014,maoGenerationComprehensionUnambiguous2016,yu_mattnet_2018,deng_transvg_2021,daiSimVGSimpleFramework} to multiple and absent referents.
GRES~\cite{liu_gres_2023} and GREC~\cite{he_grec_2023} formalize this setting for segmentation and box grounding, while GREx~\cite{dingGRExGeneralizedReferring2026a} also incorporates referring expression generation.
General-purpose segmentation frameworks, such as Mask R-CNN~\cite{he_maskrcnn_2017} and Mask2Former~\cite{chengMaskedattentionMaskTransformer2022a}, provide complementary approaches for instance-level and
pixel-level mask prediction. Specialized methods address target coverage and absence through structured decoding: CoHD~\cite{luoCoHDCountingAwareHierarchical2025} combines hierarchical mask decoding with count- and category-level supervision, whereas PropVG~\cite{daiPropVGEndtoEndProposalDriven} integrates proposal generation, referent scoring, and multi-granularity target discrimination.
Complementing these strategies, InstanceVG~\cite{daiImprovingGeneralizedVisual2026} uses instance queries to jointly predict bounding boxes and masks, promoting consistency between the two representations of each referent. HieA2G~\cite{wang_hiea2g_2025} combines hierarchical cross-modal alignment with adaptive target counting.

\noindent\textbf{MLLM-Based Visual Grounding.}
Multimodal large language models formulate grounding as autoregressive prediction of coordinates or location tokens~\cite{zhanEGMEfficientVisual2026}, as exemplified by Shikra~\cite{chenShikraUnleashingMultimodal2023}, Kosmos-2~\cite{pengGROUNDINGMULTIMODALLARGE2024a}, and the Qwen-VL family~\cite{bai_qwen25vl_2025,bai_qwen3vl_2025}.
Region-aware approaches~\cite{wuFLMMGroundingFrozen}, including Ferret~\cite{youFERRETREFERGROUND2024a}, RegionGPT~\cite{guoRegionGPTRegionUnderstanding2024}, and Osprey~\cite{yuanOspreyPixelUnderstanding2024}, enhance localized visual understanding through region-level representations.
Complementary work explores explicit spatial supervision~\cite{wuSpatialAwareTokenWeakly2023}: AMC~\cite{yangImprovingVisualGrounding2023} aligns gradient-based explanations with region annotations to improve visual grounding.
Output designs provide another direction~\cite{laiLISAReasoningSegmentation2024}: GSVA~\cite{xia_gsva_2024} uses segmentation and rejection tokens to handle multiple and absent referents, while GETok~\cite{renGroundingEverythingTokens} represents locations through learnable grid and offset tokens.
Unlike approaches that align gradient-based explanation maps,
ReSAG supervises the probability mass assigned to each target
in a shared pre-answer spatial distribution.
It retains the original language vocabulary for coordinate
generation and introduces action tokens at box boundaries
to decide whether to generate another box or terminate the sequence.

\section{The AgroVG+ Dataset}

\label{sec:agrovg_plus}

\label{sec:dataset}

AgroVG+ builds on AgroVG~\cite{liAgroVGLargeScaleMultiSource2026} through renewed image-quality selection and expanded sampling from the original data sources. Eight agricultural-domain experts manually write and review all referring expressions and verify their corresponding target labels. Together, these efforts establish a resource for generalized agricultural grounding that connects language descriptions with their target sets, including expressions with no matching referent.

\subsection{Dataset Construction}
\label{sec:agrovg_plus_construction}
\label{sec:dataset_construction}

AgroVG+ combines images reselected from AgroVG based on visual quality with additional samples from the original data sources.
The resulting collection spans six scene families: crops and weeds, fruits, wheat heads, pests, plant diseases, and tree canopies.
It comprises 16 source subsets, with source identities and instance associations preserved throughout curation.

The annotation team comprised eight experts with interdisciplinary backgrounds in agriculture and computer science and prior experience in dataset development.
Before annotation began, all participants received standardized training on the annotation workflow and the use of the annotation platform.
The annotators wrote every referring expression from image content, using visible appearance, relative position, and relationships among objects to specify individual targets or selected target groups.
For each expression, they manually checked whether the referring cues were supported by the image and whether the associated boxes matched the intended target set.
Ambiguous expressions and incorrect target associations were revised or removed, and identified annotation errors were corrected.

Each example associates an image $I$ and an expression $q$ with a target set $\mathcal{B}=\{b_i\}_{i=1}^{n}$, where $n=0$, $n=1$, and $n>1$ correspond to target-absent, single-target, and multi-target queries, respectively.
English and Chinese expressions share the same target set and constitute one bilingual record.
Query-specific answers are stored separately from the image-level annotation inventory, since only the instances satisfying the expression belong to its answer.

Target-absent queries require verifying that no instance satisfies the complete description.
The annotators wrote these queries using plausible combinations of scene-specific attributes and relations, and inspected the whole image to confirm the empty target set.
Missing source annotations alone were not treated as evidence of absence.
An empty answer therefore denotes the absence of a matching referent rather than the absence of agricultural objects in the scene.

\subsection{Dataset Statistics}

\label{sec:agrovg_plus_statistics}

\label{sec:dataset_organization}

AgroVG+ contains 7,404 images and 10,081 bilingual query records, comprising 7,318 single-target, 1,870 multi-target, and 893 target-absent queries. Figure~\ref{fig:agrovg_plus_statistics}(a,b) summarizes the source composition and query regimes across the six scene families. All three query regimes are represented in each family, supporting evaluation of instance localization and target-absence recognition across different agricultural contexts.

The collection also captures substantial variation in instance density and target scale. Figure~\ref{fig:agrovg_plus_statistics}(c) shows the distributions of annotated instances per image, whose medians range from two for pests to 29 for wheat heads. These counts describe the available source annotations, whereas Figure~\ref{fig:agrovg_plus_statistics}(d) characterizes the boxes actually referred to by queries. The latter comprises 12,261 target occurrences, including repeated references across queries, with relative areas spanning several orders of magnitude. These variations require models to distinguish relevant instances within crowded scenes and localize targets at different scales.

Dataset splits follow documented image-association groups, so images within the same group and all their associated expressions remain in the same partition. Source-specific split constraints are preserved where applicable. An internal validation set is held out from the training pool. Images, query records, and recorded association groups are disjoint across the training, validation, and test splits. Table~\ref{tab:agrovg_plus_splits} summarizes image counts and the distribution of queries across target states and splits. Further dataset statistics, annotation protocols, and quality-control procedures are provided in Supplementary Sec.~\ref{sec:sup_data}.

\begin{table}[t]
    \centering
    \small
    \setlength{\tabcolsep}{4pt}
    \caption{\textbf{Dataset statistics and split distribution of AgroVG+.}
    Single, Multi, and Absent count queries, with each bilingual record counted once.}
    \label{tab:agrovg_plus_splits}
    \begin{tabular}{@{}lrrrrr@{}}
        \toprule
        Split & Images & Single & Multi & Absent & Queries \\
        \midrule
        Train & 4,724 & 4,627 & 1,150 & 576 & 6,353 \\
        Val.  &   459 &   525 &   129 &  67 &   721 \\
        Test  & 2,221 & 2,166 &   591 & 250 & 3,007 \\
        \midrule
        Total & 7,404 & 7,318 & 1,870 & 893 & 10,081 \\
        \bottomrule
    \end{tabular}
\end{table}

\section{Method}
\label{sec:method}
\begin{figure*}[t]
    \centering
    \includegraphics[width=\linewidth]{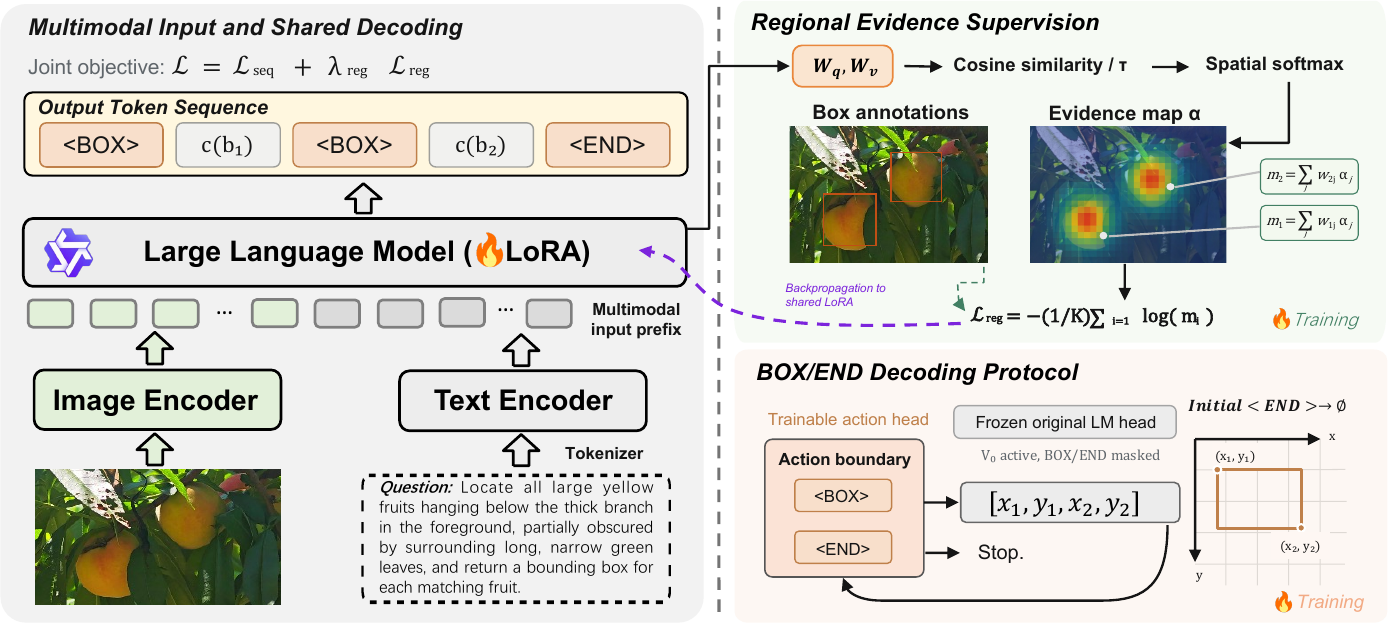}
    \caption{\textbf{Regional supervision guides action generation through a shared decoder.}
    \emph{Left:} Image and query embeddings form the multimodal prefix; the text branch denotes tokenization and embedding lookup.
    The joint objective combines teacher-forced sequence loss and regional loss.
    \emph{Upper right:} During training, projections of final-layer prefix states yield spatial evidence $\boldsymbol{\alpha}$.
    Box annotations supervise each target's evidence mass $m_i=\sum_j w_{ij}\alpha_j$; the purple dashed arrow denotes backpropagation to the shared LoRA adapters.
    This evidence branch is used only during training.
    \emph{Lower right:} At each action boundary, \texttt{<BOX>} starts coordinate generation with the frozen language head, while \texttt{<END>} terminates the sequence; an initial \texttt{<END>} returns $\varnothing$.}
    \label{fig:resag_overview}
\end{figure*}

Given an image $I$ and a referring expression $q$, ReSAG predicts a set of matching bounding boxes that may be empty.
As shown in Fig.~\ref{fig:resag_overview}, a shared autoregressive decoder provides the representations for regional evidence supervision and generates the output sequence under the BOX/END protocol.
During training, annotated boxes supervise a spatial evidence distribution computed from the image--expression prefix, while action and coordinate tokens provide sequence supervision.
These objectives jointly adapt the decoder, which generates boxes and termination actions from its own output history at inference.

\subsection{Regional Evidence Supervision}
\label{sec:regional_supervision}

To provide a direct spatial training signal for every annotated referent, we supervise a shared evidence distribution over image locations.
Let $\mathbf{h}^{\mathrm{pre}}\in\mathbb{R}^{d}$ denote the decoder hidden state at the final position of the image--expression prefix, immediately before answer generation.
Let $\mathbf{v}_j\in\mathbb{R}^{d}$ denote the hidden state at image-token position $j$.
Both representations are taken from the final language decoder layer, and the $J$ image-token positions correspond to spatial cells $\{S_j\}_{j=1}^{J}$ on the merged visual grid. The $H_g\times W_g$ merged grid partitions the original $W\times H$ image into cells of size $(W/W_g)\times(H/H_g)$,
indexed in row-major order.

We project the prefix state and image-position states into a common space using learnable matrices $W_q,W_v\in\mathbb{R}^{p\times d}$.
Their similarities define the spatial evidence distribution:
\begin{equation}
    \begin{aligned}
        s_j &=
        \frac{\operatorname{cos}
        (W_q\mathbf{h}^{\mathrm{pre}},W_v\mathbf{v}_j)}
        {\tau}, \\
        \alpha_j &=
        \frac{\exp(s_j)}
        {\sum_{\ell=1}^{J}\exp(s_\ell)},
    \end{aligned}
    \label{eq:regional_evidence}
\end{equation}
where $\operatorname{cos}$ denotes cosine similarity and $\tau>0$ is a temperature.
The resulting distribution assigns evidence mass to image locations conditioned on the input prefix.
Under causal attention, these prefix representations have no access to the answer coordinates.

For an annotated target set $\mathcal{B}=\{b_i\}_{i=1}^{K}$, each box defines a separate spatial constraint on this distribution.
We assign cell $S_j$ a weight according to the fraction of its area covered by box $b_i$:
\begin{equation}
    w_{ij}=\frac{|b_i\cap S_j|}{|S_j|},
    \label{eq:regional_weights}
\end{equation}
where $|\cdot|$ denotes area in the image coordinate system.
These weights preserve partial coverage at box boundaries.
The weighted evidence mass associated with target $i$ is
$\sum_{j=1}^{J}w_{ij}\alpha_j$, yielding the regional objective
\begin{equation}
    \mathcal{L}_{\mathrm{reg}}=
    \begin{cases}
        -\dfrac{1}{K}\displaystyle\sum_{i=1}^{K}
        \log\!\left(
        \displaystyle\sum_{j=1}^{J}w_{ij}\alpha_j
        \right), & K>0, \\[4pt]
        0, & K=0.
    \end{cases}
    \label{eq:regional_loss}
\end{equation}
Each target contributes one equally weighted loss term to the same prefix evidence distribution.
Concentrating evidence on a prominent target therefore leaves a penalty for other targets with low regional evidence mass.
This objective uses the available box annotations to encourage spatial support across the target set without requiring segmentation masks.
For target-absent queries, the regional loss is zero, and supervision is supplied by the output sequence.

\subsection{BOX/END Action Generation}
\label{sec:action_generation}

To make box continuation and response termination explicit, we represent the target set using two atomic action tokens, \texttt{<BOX>} and \texttt{<END>}.
We order the target boxes lexicographically by their normalized coordinates to obtain a deterministic training sequence:
\begin{equation}
    Y=\texttt{<BOX>}\,c(b_1)\;\cdots\;
    \texttt{<BOX>}\,c(b_K)\,\texttt{<END>},
    \label{eq:action_sequence}
\end{equation}
where $c(b_i)$ denotes the tokenized coordinate string
$[x_1,y_1,x_2,y_2]$, with $(x_1,y_1)$ and $(x_2,y_2)$ representing the top-left and bottom-right corners, respectively.
Horizontal and vertical coordinates are scaled to $[0,1000]$ using the image width and height, respectively.
For an empty target set, the complete sequence is \texttt{<END>}.
Coordinates and punctuation use the original language vocabulary, while the two action tokens have learnable input embeddings and output vectors.

Generation alternates between an action boundary and a coordinate body.
The answer begins at an action boundary, where the decoder selects \texttt{<BOX>} to enter coordinate generation or \texttt{<END>} to terminate.
Within a coordinate body, both action tokens are masked.
Once the decoded body ends with the closing bracket \texttt{]}, generation returns to an action boundary.
Let $\mathcal{A}=\{\texttt{<BOX>},\texttt{<END>}\}$ denote the action vocabulary and $\mathcal{V}_0$ the original tokenizer vocabulary.
The active vocabulary at step $t$ is
\begin{equation}
    \mathcal{V}_t=
    \begin{cases}
        \mathcal{A}, & \text{at an action boundary},\\
        \mathcal{V}_0, & \text{within a coordinate body}.
    \end{cases}
    \label{eq:active_vocabulary}
\end{equation}

Let $\mathbf{h}_t$ denote the decoder hidden state predicting token $y_t$, with $\mathbf{h}_1=\mathbf{h}^{\mathrm{pre}}$.
For an action token $v\in\mathcal{A}$, its logit is
$\ell_t(v)=\mathbf{a}_v^\top\mathbf{h}_t$,
where $\mathbf{a}_v$ is its learnable output vector.
Coordinate-body logits are produced by the pretrained language head.
Writing $p_t(v)=p_\theta(y_t=v\mid I,q,y_{<t})$, we normalize over the active vocabulary:
\begin{equation}
    p_t(v)=
    \frac{\exp(\ell_t(v))}
    {\sum_{u\in\mathcal{V}_t}\exp(\ell_t(u))},
    \qquad v\in\mathcal{V}_t,
    \label{eq:action_probability}
\end{equation}
with $p_t(v)=0$ outside $\mathcal{V}_t$.
Thus, continuation and termination compete through a two-way normalization at each boundary, while coordinate generation retains the original vocabulary.

The same vocabulary constraints apply during training and inference.
Action and coordinate tokens are supervised together through the sequence objective described next.

\begin{figure*}[t]
    \centering
    \includegraphics[width=\textwidth]{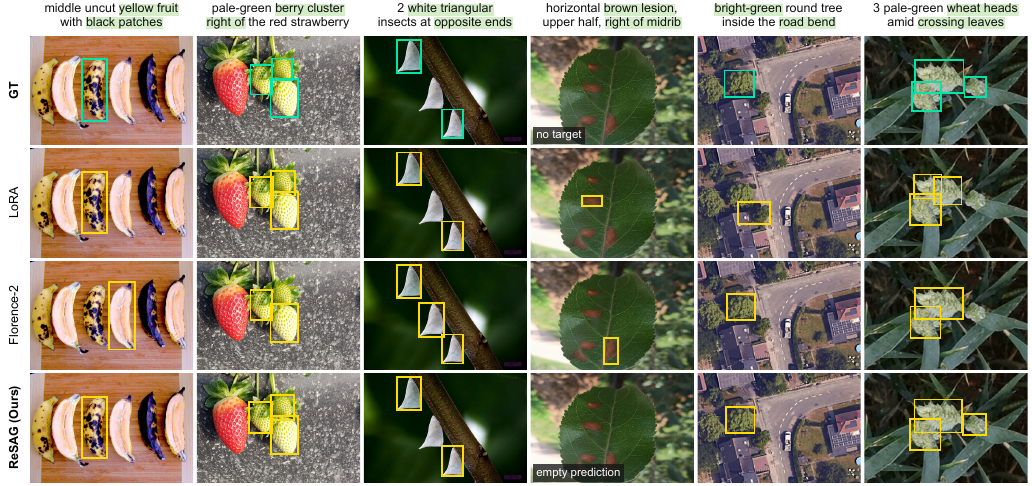}
    \caption{\textbf{Qualitative comparison on AgroVG+.}
    Selected examples illustrate instance confusion, missed or extra
    targets, and target absence across agricultural scenes.
    Green boxes denote ground truth; yellow boxes denote predictions.
    Queries are abbreviated.}
    \label{fig:resag_prediction_comparison}
\end{figure*}

\subsection{Joint Training and Inference}
\label{sec:training_inference}

We jointly optimize regional evidence learning and action-based box generation through the shared decoder.
For a target sequence $Y=(y_1,\ldots,y_T)$, we use the constrained probabilities defined in Eq.~\eqref{eq:action_probability} to compute
\begin{equation}
    \mathcal{L}_{\mathrm{seq}}
    =-\frac{1}{T}\sum_{t=1}^{T}
    \log p_\theta(y_t\mid I,q,y_{<t}).
    \label{eq:sequence_loss}
\end{equation}
The token average includes both action tokens and all coordinate-body tokens, while input-prefix tokens are excluded from the sequence loss.
Training uses teacher forcing, so each sequence prediction conditions on preceding ground-truth answer tokens.
The regional evidence distribution uses only input-prefix states, which cannot attend to answer tokens under causal attention.

For each query, the complete objective is
\begin{equation}
    \mathcal{L}
    =\mathcal{L}_{\mathrm{seq}}
    +\lambda_{\mathrm{reg}}\mathcal{L}_{\mathrm{reg}},
    \label{eq:joint_loss}
\end{equation}
where $\lambda_{\mathrm{reg}}$ controls the contribution of regional supervision.
Per-query objectives are averaged with equal weights within each optimization update.
Continuation, termination, and coordinate prediction are all learned through $\mathcal{L}_{\mathrm{seq}}$.

In our implementation, the trainable parameters $\theta$ comprise language LoRA adapters~\cite{huLORALOWRANKADAPTATION2022}, the regional projections $W_q$ and $W_v$, and the input embeddings and output vectors of the two action tokens.
The original vision encoder, visual mergers, and pretrained language-model parameters remain frozen.
Both objectives update the shared language adapters, allowing regional supervision to influence generation through the decoder parameters.
The evidence map serves only as an auxiliary training signal, while inference uses the adapted decoder.

At inference, we perform greedy decoding with a key/value cache and the vocabulary constraints described in Sec.~\ref{sec:action_generation}.
Emitting \texttt{<END>} terminates the response, with immediate termination producing an empty prediction.
The entire response is counted as incorrect if it reaches the
token budget without \texttt{<END>}, violates the BOX/END format,
or contains an invalid box.
Valid boxes contain four finite coordinates satisfying
$0\le x_1<x_2\le1000$ and $0\le y_1<y_2\le1000$.
No partial recovery, coordinate repair, or duplicate suppression
is applied. Further training and decoding details are provided
in Supplementary Sec.~\ref{sec:sup_impl}.

\section{Experiments}
\label{sec:experiments}

\begin{table*}[t]
    \centering
    \caption{\textbf{Comparison on AgroVG+.}
    Best and second-best distinct scores are bold
    and underlined, respectively.}
    \label{tab:agrovg_plus_main_comparison}

    \small
    \setlength{\tabcolsep}{3.5pt}
    \renewcommand{\arraystretch}{1.08}
    \begin{tabular}{@{}lc@{\hspace{4pt}}ccccccc@{}}
        \toprule
        \multirow[c]{2}{*}{\textbf{Method}}
        & \multirow[c]{2}{*}{\textbf{Adaptation}}
        & \multirow[c]{2}{*}{\shortstack[c]{
            \textbf{Trainable}\\
            \textbf{Params. (M)}}}
        & \multicolumn{2}{c}{\textbf{Single-target}}
        & \multicolumn{2}{c}{\textbf{Multi-target}}
        & \multirow[c]{2}{*}{\textbf{N-acc}}
        & \textbf{Overall} \\
        \cmidrule(lr){4-5}
        \cmidrule(lr){6-7}
        \cmidrule(lr){9-9}
        & & & @0.50 & @0.75 & @0.50 & @0.75
        & & $\mathrm{Pr}@(F_1=1)$ \\
        \midrule

        \multicolumn{9}{@{}l}{
            \textit{Detection and foundation models}} \\

        MDETR-GREC~\cite{kamathMDETRModulatedDetection2021a,he_grec_2023}
        & Full FT & 185.16
        & 39.24 & 30.06 & 16.07 & 12.35
        & 9.20 & 32.19 \\

        Grounding DINO-T~\cite{liu_grounding_2025}
        & Full FT & 172.25
        & 62.00 & 54.52 & 30.63 & 25.38
        & 40.40 & 54.04 \\

        Florence-2-large-ft~\cite{xiao_florence2_2024}
        & Full FT & 770.17
        & \underline{77.33} & \textbf{68.84}
        & 58.88 & \textbf{50.42}
        & 76.40 & 73.63 \\
        \midrule

        \multicolumn{9}{@{}l}{
            \textit{LLM-based grounding models}} \\

        Shikra-7B~\cite{chenShikraUnleashingMultimodal2023}
        & LoRA & 79.95
        & 45.61 & 19.11 & 26.06 & 5.92
        & 72.40 & 44.00 \\

        Ferret-7B~\cite{youFERRETREFERGROUND2024a}
        & None & 0.00
        & 19.48 & 9.46 & 0.51 & 0.17
        & 0.00 & 14.13 \\

        Ferret-7B~\cite{youFERRETREFERGROUND2024a}
        & LoRA & 79.95
        & 54.52 & 26.96 & 34.52 & 12.18
        & 69.60 & 51.85 \\

        Qwen3-VL-8B~\cite{bai_qwen3vl_2025}
        & None & 0.00
        & 51.89 & 37.12 & 32.32 & 20.98
        & 59.20 & 48.65 \\

        Qwen3-VL-8B~\cite{bai_qwen3vl_2025}
        & LoRA & 87.29
        & 76.45 & 55.96
        & \underline{67.51} & \underline{45.69}
        & \underline{82.80} & \underline{75.22} \\

        InternVL3.5-8B~\cite{wangInternVL35AdvancingOpenSource2025}
        & LoRA & 87.29
        & 75.07 & 55.72 & 65.99 & 44.50
        & 81.60 & 73.83 \\
        \midrule

        \textbf{ReSAG (Ours)}
        & LoRA$^{*}$ & 88.36
        & \textbf{77.89} & \underline{59.14}
        & \textbf{71.40} & \textbf{50.42}
        & \textbf{83.60} & \textbf{77.09} \\
        \bottomrule
    \end{tabular}

    \par\smallskip
    \parbox{0.97\textwidth}{\footnotesize
    $^{*}$LoRA with trainable regional projections and BOX/END
    parameters, introducing 1.06M additional trainable parameters.}
\end{table*}

\subsection{Settings}
\label{sec:settings}

All experiments are conducted on AgroVG+ using the training,
validation, and test splits described in Sec.~\ref{sec:dataset}.
We use only the English expressions for training, validation, and testing.
The Chinese counterparts are retained in the dataset but are not used in any experiment.

\noindent\textbf{Evaluation metrics.}
We report exact-set accuracy (ESA) at IoU
0.50 and 0.75 for single-target and multi-target queries.
At each threshold, ESA counts a query as correct only when
maximum-cardinality one-to-one matching pairs every predicted
box with a ground-truth box and leaves no box unmatched.
N-acc measures the fraction of target-absent queries correctly
predicted as empty.
Overall, denoted $\mathrm{Pr}@(F_1=1)$, measures exact-set
accuracy across all queries at IoU 0.50, including target-absent
queries with valid empty predictions.
Invalid or missing outputs are counted as incorrect;
scores are percentages. Supplementary Sec.~\ref{sec:sup_eval} details the
matching and output-validation rules.

\noindent\textbf{Compared methods.}
We compare ReSAG with two groups of models
(Table~\ref{tab:agrovg_plus_main_comparison}).
Detection and foundation models include
MDETR-GREC~\cite{kamathMDETRModulatedDetection2021a,he_grec_2023},
Grounding DINO-T~\cite{liu_grounding_2025},
and Florence-2-large-ft~\cite{xiao_florence2_2024},
adapted through full-parameter fine-tuning.
LLM-based grounding models include
Shikra-7B~\cite{chenShikraUnleashingMultimodal2023},
Ferret-7B~\cite{youFERRETREFERGROUND2024a},
Qwen3-VL-8B~\cite{bai_qwen3vl_2025},
and InternVL3.5-8B~\cite{wangInternVL35AdvancingOpenSource2025},
adapted through language LoRA.
Additional zero-shot results are reported for Ferret-7B
and Qwen3-VL-8B.
LoRA-adapted Qwen3-VL-8B serves as the same-backbone
baseline for ReSAG.

\noindent\textbf{Implementation details.}
We initialized ReSAG from Qwen3-VL-8B-Instruct and applied
LoRA to the attention and feed-forward projections of the
language decoder, with rank 32, $\alpha_{\mathrm{LoRA}}=64$,
and dropout 0.05.
The regional settings were $p=128$, $\tau=0.1$, and
$\lambda_{\mathrm{reg}}=0.1$.
Only the LoRA adapters, regional projections, and BOX/END
parameters were optimized; all pretrained weights remained frozen.
Training used BF16 on two NVIDIA A100 GPUs for six epochs
(4,770 updates), with an effective batch size of eight queries.
We used AdamW with a learning rate of $5\times10^{-5}$ and
weight decay 0.01, together with cosine decay and
144 warmup updates.
The image processor used a maximum pixel budget of 524,288,
and greedy decoding generated at most 4,096 tokens.
Coordinate values in the training targets were rounded to
three decimal places.
The input embeddings and output vectors of \texttt{<BOX>}
and \texttt{<END>} were initialized from the pretrained
\texttt{[} and end-of-response token rows, respectively.
We selected the checkpoint with the best validation performance
on the primary evaluation metric and used it for final
evaluation on the test set.

\begin{figure*}[t]
    \centering
    \includegraphics[width=\textwidth]{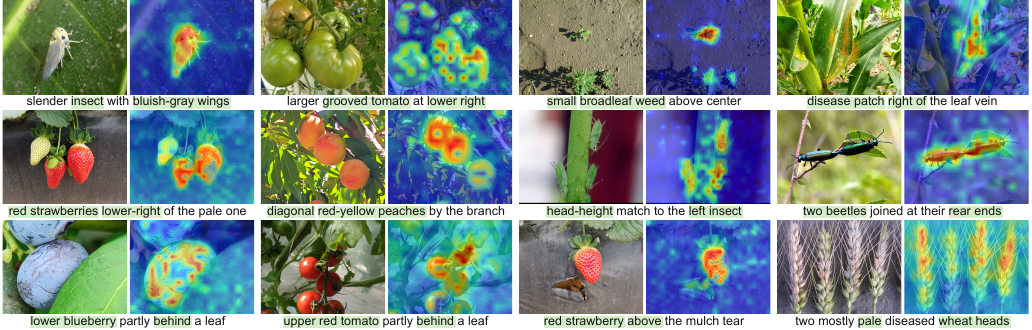}
    \caption{\textbf{Regional evidence learned by ReSAG.}
    Each pair shows an image crop (left) and the query-conditioned regional evidence computed before answer generation (right).
    Labels summarize the corresponding queries.}
    \label{fig:resag_regional_evidence}
\end{figure*}

\subsection{Results}
\label{sec:results}
ReSAG achieves the highest overall exact-set accuracy of
77.09\% among the evaluated methods on AgroVG+
(Table~\ref{tab:agrovg_plus_main_comparison}).
The comparison with Florence-2 illustrates the distinction
between single-target localization and complete-set recovery.
Florence-2 achieves the highest single-target ESA@0.75,
but its multi-target ESA@0.50 is 58.88\%, compared with
71.40\% for ReSAG.
ReSAG also matches Florence-2 on multi-target ESA@0.75
and achieves the highest target-absence accuracy.
These results show that ReSAG performs well on both
recovering multiple matching instances and identifying
queries with no valid target, key requirements
of generalized agricultural grounding.

Compared with the reported LoRA baseline using
the same Qwen3-VL-8B backbone, ReSAG achieves higher scores across all
evaluation metrics.
The gains are larger for multi-target queries, with
improvements of 3.89 and 4.74 percentage points in ESA
at IoU thresholds of 0.50 and 0.75, respectively.
ReSAG introduces approximately 1.06M additional trainable
parameters through its regional projections and BOX/END
parameters, while keeping all pretrained weights frozen.
The component comparisons in
Sec.~\ref{sec:ablation_analysis} further examine the
individual and combined effects of these two designs.

\noindent\textbf{Qualitative comparison.}
Figure~\ref{fig:resag_prediction_comparison} presents
selected predictions from LoRA, Florence-2, and ReSAG.
ReSAG recovers all three targets in the strawberry and
wheat examples, whereas Florence-2 misses one target
in each.
For wheat, LoRA outputs three boxes but selects the wrong
instances, showing that correct counts alone are insufficient.
For the target-absent lesion query, ReSAG returns an
empty prediction, while both baselines produce an
incorrect box.
These examples illustrate differences in target selection,
instance coverage, and target-absence recognition.

\subsection{Ablation Studies and Analysis}
\label{sec:ablation_analysis}

\noindent\textbf{Effect of regional supervision.}
Adding regional supervision improves Overall from 75.22\% to
76.06\%, while increasing all four ESA metrics
(Table~\ref{tab:agrovg_plus_component_ablation}).
At IoU 0.75, the gains reach 2.35 percentage points for
single-target queries and 4.23 points for multi-target queries.
When added to the BOX/END-only variant, regional supervision
further raises Overall from 75.69\% to 77.09\% and multi-target
ESA@0.75 from 47.55\% to 50.42\%.
These improvements are consistent with the regional objective,
which assigns an equally weighted term to each
annotated target to encourage evidence coverage across the
complete target set.

\noindent\textbf{Effect of the BOX/END protocol.}
Adding the BOX/END protocol improves Overall from 75.22\% to
75.69\% over the LoRA baseline
(Table~\ref{tab:agrovg_plus_component_ablation}).
Its contribution is larger when combined with regional
supervision: Overall increases from 76.06\% for the Region
variant to 77.09\% for ReSAG, a gain of 1.03 percentage points.
The joint configuration achieves the highest Overall among
the four variants, exceeding the LoRA baseline by
1.86 percentage points.
These gains support combining regional supervision and BOX/END
decisions in a shared decoder for complete-set prediction.

\noindent\textbf{Regional evidence analysis.}
Figure~\ref{fig:resag_regional_evidence} visualizes the query-conditioned spatial evidence computed before answer generation. For single-target queries, high responses concentrate around the slender insect and small weed. The strawberry and beetle examples show elevated evidence across multiple matching instances, while the partially occluded blueberry receives strong responses on its
visible surface. These patterns are consistent with the regional objective,
which gives each annotated instance an equally weighted contribution to learning the shared evidence distribution. Some responses also extend to neighboring objects and vegetation. The regional loss supervises evidence mass within boxes,
without requiring the maps to follow precise object boundaries. Additional analyses, visualizations, and failure cases are presented in Supplementary Sec.~\ref{sec:sup_analysis}.

\begin{table}[t]
    \centering
    \caption{\textbf{Component ablation on AgroVG+.}
    $\Delta$ denotes percentage-point gains over LoRA, computed before rounding.
    Best and second-best scores are bold and underlined, respectively.}
    \label{tab:agrovg_plus_component_ablation}

    \footnotesize
    \setlength{\tabcolsep}{3pt}
    \renewcommand{\arraystretch}{1.17}
    \begin{tabular}{lccccc}
        \toprule
        \multirow{2}{*}{\textbf{Variant}}
        & \multicolumn{2}{c}{\textbf{Single-target}}
        & \multicolumn{2}{c}{\textbf{Multi-target}}
        & \multirow{2}{*}{\textbf{Overall}} \\
        \cmidrule(lr){2-3}
        \cmidrule(lr){4-5}
        & @0.50 & @0.75 & @0.50 & @0.75 & \\
        \midrule

        LoRA baseline
        & 76.45 & 55.96
        & 67.51 & 45.69
        & 75.22 \\

        \rowcolor{ablationgray}
        + Region
        & 77.05 & \underline{58.31}
        & 70.05 & \underline{49.92}
        & \underline{76.06} \\

        \hspace{0.5em}{\scriptsize $\Delta$ vs. LoRA}
        & {\scriptsize $+0.60$}
        & {\scriptsize $+2.35$}
        & {\scriptsize $+2.54$}
        & {\scriptsize $+4.23$}
        & {\scriptsize $+0.83$} \\

        \rowcolor{ablationgray}
        + BOX/END
        & \underline{77.24} & 57.99
        & \underline{70.22} & 47.55
        & 75.69 \\

        \hspace{0.5em}{\scriptsize $\Delta$ vs. LoRA}
        & {\scriptsize $+0.78$}
        & {\scriptsize $+2.03$}
        & {\scriptsize $+2.71$}
        & {\scriptsize $+1.86$}
        & {\scriptsize $+0.47$} \\

        \rowcolor{ablationgray}
        \textbf{ReSAG}
        & \textbf{77.89} & \textbf{59.14}
        & \textbf{71.40} & \textbf{50.42}
        & \textbf{77.09} \\

        \hspace{0.5em}{\scriptsize $\Delta$ vs. LoRA}
        & {\scriptsize $+1.43$}
        & {\scriptsize $+3.19$}
        & {\scriptsize $+3.89$}
        & {\scriptsize $+4.74$}
        & {\scriptsize $+1.86$} \\
        \bottomrule
    \end{tabular}
\end{table}

\section{Conclusion}
\label{sec:conclusion}

We introduced AgroVG+, a dataset for generalized agricultural
grounding with expert-written referring expressions based on scene
content and manually reviewed target sets.
It covers single-target, multi-target, and target-absent
queries.
We also presented ReSAG, which uses box annotations to
supervise spatial evidence before answer generation and
to learn explicit BOX/END decisions.
The regional and sequence objectives are jointly optimized
through a shared language decoder to learn where the
targets are and when to continue or stop generating boxes.
Experiments on AgroVG+ show that ReSAG achieves the highest
overall exact-set accuracy among the evaluated methods,
while component ablations show that combining regional
supervision with BOX/END outperforms either component alone.
\section*{LLM Usage Statement}

\noindent We used ChatGPT (GPT-6 Astra, OpenAI) to help organize
and format research data, implement the proposed method, debug
code, and resolve software dependency conflicts and kernel-related
issues.
The tool also assisted with preparing visualizations, refining
the presentation of figures and tables, and polishing
author-written text.
It was not used to determine the method design or experimental
settings, which were decided by the authors.
All AI-assisted outputs were reviewed and validated by the
authors, and all experimental results were interpreted by
the authors.
The authors take full responsibility for the final content
of this work.

\pdfbookmark[0]{References}{references}
{
    \small
    \bibliographystyle{ieeenat_fullname}
    \bibliography{main,source_datasets}
}

\twocolumn[{%
  \pdfbookmark[0]{Supplementary Material}{supplementary}
  \centering\Large\bfseries Supplementary Material\par
  \vspace{1em}
}]
\appendix
\raggedbottom
\setlength{\dblfloatsep}{8pt plus 2pt minus 2pt}

\setcounter{table}{0}
\setcounter{figure}{0}
\setcounter{equation}{0}
\renewcommand{\thetable}{S\arabic{table}}
\renewcommand{\thefigure}{S\arabic{figure}}
\renewcommand{\theequation}{S\arabic{equation}}
\renewcommand{\theHtable}{\thetable}
\renewcommand{\theHfigure}{\thefigure}
\renewcommand{\theHequation}{\theequation}

\newcommand{\BOX}{\texttt{<BOX>}}
\newcommand{\END}{\texttt{<END>}}
\newcommand{\ind}[1]{\mathbf{1}\!\left[#1\right]}
\newcolumntype{L}{>{\raggedright\arraybackslash}X}
\newcommand{\AgroVGReleaseLink}[2]{%
  \if\relax\detokenize{#2}\relax#1\else\href{#2}{#1}\fi}

\noindent This supplement is organized into four sections. Section~\ref{sec:sup_data} documents AgroVG+ and its annotations. Section~\ref{sec:sup_impl} describes training and decoding, and Sec.~\ref{sec:sup_eval} specifies the evaluation protocol. Section~\ref{sec:sup_analysis} outlines additional analyses and visualizations, with the scope of interpretation discussed at the end of the failure-case analysis in Sec.~\ref{sec:sup_failures}.

\section{Additional Dataset Documentation}
\label{sec:sup_data}

\subsection{Sources and annotation units}

\begin{figure*}[!tp]
\centering
\includegraphics[width=\textwidth]{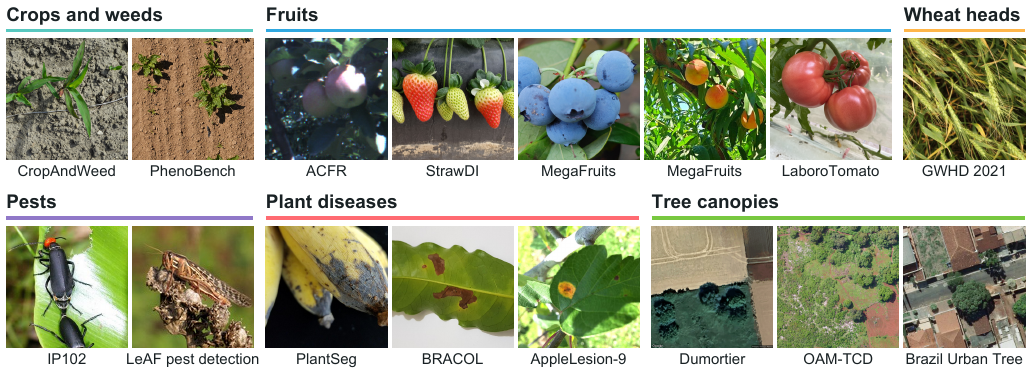}
\caption{\textbf{Visual diversity across AgroVG+ sources.}
We group the 16 subsets by scene family, showing one training image from each.}
\label{fig:sup_source_gallery}
\end{figure*}

We constructed AgroVG+ by rescreening images from
AgroVG~\cite{liAgroVGLargeScaleMultiSource2026} for quality and
sampling additional images from the original datasets.
AgroVG+ contains 7,404 images and 10,081 bilingual query records
from 16 source subsets across six scene families.
Figure~\ref{fig:sup_source_gallery} shows one training image from each subset.
Table~\ref{tab:sup_sources} reports image and query counts by subset,
with separate counts for the blueberry and peach subsets of MegaFruits.
For each source dataset, Table~\ref{tab:sup_source_annotations}
cites the relevant papers and data releases and describes how the
image-level boxes were obtained.

We count an English referring expression $q$ and its Chinese
counterpart as one bilingual query record for an image $I$.
Their shared target set $\mathcal{B}$ consists of the boxes of all
annotated instances in $I$ that satisfy the full expression.
Different expressions for the same image can therefore yield a single
box, multiple boxes, or an empty set.
Only the English expressions are used for training, validation, and testing.

\begin{table*}[!tp]
\centering
\small
\setlength{\tabcolsep}{5pt}
\caption{\textbf{Source composition of AgroVG+.}}
\label{tab:sup_sources}
\renewcommand{\arraystretch}{0.96}
\begin{tabular*}{\textwidth}{@{\extracolsep{\fill}}llrrrrr@{}}
\toprule
\textbf{Scene family} & \textbf{Source subset} & \textbf{Images} & \textbf{Single} & \textbf{Multi} & \textbf{Absent} & \textbf{Queries} \\
\midrule
Crops and weeds & CropAndWeed & 507 & 445 & 109 & 54 & 608 \\
& PhenoBench & 886 & 749 & 280 & 86 & 1,115 \\
\addlinespace[1.5pt]
Fruits & ACFR & 587 & 506 & 92 & 24 & 622 \\
& StrawDI & 320 & 232 & 137 & 52 & 421 \\
& MegaFruits (blueberry) & 138 & 131 & 55 & 25 & 211 \\
& MegaFruits (peach) & 137 & 123 & 45 & 20 & 188 \\
& LaboroTomato & 290 & 253 & 118 & 49 & 420 \\
\addlinespace[1.5pt]
Wheat heads & GWHD 2021 & 1,049 & 1,043 & 147 & 119 & 1,309 \\
\addlinespace[1.5pt]
Pests & IP102 & 766 & 582 & 245 & 101 & 928 \\
& LeAF pest detection & 262 & 253 & 34 & 49 & 336 \\
\addlinespace[1.5pt]
Plant diseases & PlantSeg & 729 & 726 & 152 & 55 & 933 \\
& BRACOL & 252 & 193 & 96 & 60 & 349 \\
& AppleLesion & 184 & 123 & 122 & 34 & 279 \\
\addlinespace[1.5pt]
Tree canopies & Dumortier & 460 & 424 & 83 & 60 & 567 \\
& OAM-TCD & 629 & 1,296 & 113 & 37 & 1,446 \\
& Brazil Urban Tree & 208 & 239 & 42 & 68 & 349 \\
\midrule
Total & 16 source subsets & 7,404 & 7,318 & 1,870 & 893 & 10,081 \\
\bottomrule
\end{tabular*}
\end{table*}

\begin{table*}[t]
\centering
\small
\caption{\textbf{Source annotations and access for AgroVG+.}
Year refers to the dataset paper or primary data record.
Det., seg., and cls. denote detection, segmentation, and classification.
For masks, polygons, and circles, $\to$ denotes conversion to an enclosing box.}
\label{tab:sup_source_annotations}
\fontsize{7.25}{9}\selectfont
\setlength{\tabcolsep}{1.2pt}
\renewcommand{\arraystretch}{1.20}
\begin{tabular*}{\textwidth}{@{\extracolsep{\fill}}lcllll@{}}
\toprule
\textbf{Source dataset} & \textbf{Year} & \textbf{Family}
& \textbf{Original task and labels} & \textbf{Box provenance}
& \textbf{Source URL} \\
\midrule
CropAndWeed~\cite{steininger_cropandweed_2023}
& 2023 & Crops and weeds
& \scalebox{0.91}[1]{Det./seg./localization (boxes, semantic masks, stem points)}
& Native plant boxes
& \href{https://github.com/cropandweed/cropandweed-dataset}{GitHub} \\
PhenoBench~\cite{weyler_phenobench_2024}
& 2024 & Crops and weeds
& Seg. (semantic and plant/leaf instance masks)
& Plant instance masks $\to$~boxes
& \href{https://www.phenobench.org/dataset.html}{PhenoBench} \\
\addlinespace[2pt]
ACFR~\cite{bargoti_orchard_2017}
& 2017 & Fruits
& Det. (apple circles, almond/mango boxes)
& \scalebox{0.91}[1]{Apple circles $\to$~boxes\enspace Almond \texttt{xywh}$\to$\texttt{xyxy}}
& \href{https://data.acfr.usyd.edu.au/ag/treecrops/2016-multifruit/}{ACFR} \\
StrawDI~\cite{perez_strawdi_2020}
& 2020 & Fruits
& Instance seg. (fruit masks)
& Strawberry instance masks $\to$~boxes
& \href{https://strawdi.github.io/}{StrawDI} \\
MegaFruits~\cite{wang_megafruits_2026,wang_megafruits_data_v2}
& 2026 & Fruits
& Det./seg. (visible-fruit polygons)
& Blueberry and peach polygons $\to$~boxes
& \href{https://www.kaggle.com/datasets/mmwang0/megafruits/versions/2}{Kaggle} \\
LaboroTomato~\cite{laboro_tomato_dataset}
& 2020 & Fruits
& Det./seg. (instance masks, ripeness labels)
& Orientation-corrected polygons $\to$~boxes
& \href{https://github.com/laboroai/LaboroTomato}{GitHub}\enspace \\
\addlinespace[2pt]
GWHD 2021~\cite{david_gwhd_2020,david_gwhd_2021,david_gwhd_data_v1}
& 2021 & Wheat heads
& Det. (head boxes)
& Native wheat-head boxes
& \href{https://zenodo.org/records/5092309}{Zenodo} \\
\addlinespace[2pt]
IP102~\cite{wu_ip102_2019}
& 2019 & Pests
& Cls./det. (class labels, boxes)
& Detection-subset boxes
& \href{https://github.com/xpwu95/IP102}{GitHub} \\
LeAF pest detection~\cite{sengupta_leaf_pests_2025}
& 2025 & Pests
& Cls./det. (class IDs, YOLO boxes)
& Pest boxes from annotation export
& \href{https://data.aifarms.org/view/leaf_pest_detection}{AIFARMS} \\
\addlinespace[2pt]
PlantSeg~\cite{wei_plantseg_paper_2026,wei_plantseg_coco_v7}
& 2025 & Plant diseases
& Seg. (lesion masks, COCO boxes)
& COCO lesion boxes
& \href{https://zenodo.org/records/17719108}{Zenodo} \\
BRACOL~\cite{krohling_bracol_2019,fragoso_bracol_detection_2025}
& 2019 & Plant diseases
& Symptom cls./seg. (lesion boxes in det. variant)
& BRACOL-VALIDADO v3 boxes
& \href{https://data.mendeley.com/datasets/yy2k5y8mxg/1}{Mendeley}\enspace \\
AppleLesion-9~\cite{applelesion9_source_snapshot,yang_appleleaf9_2022}
& 2026 & Plant diseases
& Det. (YOLO lesion boxes)
& Boxes from the v1 snapshot
& \href{https://data.mendeley.com/datasets/htbd8s567w/1}{Mendeley} \\
\addlinespace[2pt]
Dumortier~\cite{dumortier_crowns_2025}
& 2025 & Tree canopies
& Det. (urban/rural crown boxes)
& Native and added crown boxes
& \href{https://zenodo.org/records/15155081}{Zenodo} \\
OAM-TCD~\cite{veitch_oamtcd_2024,veitch_oamtcd_release_v1}
& 2024 & Tree canopies
& Canopy seg. (instance polygons, semantic masks)
& Per-tree COCO boxes
& \href{https://zenodo.org/records/11617167}{Zenodo} \\
Brazil Urban Tree~\cite{zamboni_urban_tree_2021}
& 2021 & Tree canopies
& Det. (crown boxes in aerial tiles)
& Campo Grande crown boxes
& \href{https://github.com/pedrozamboni/individual_urban_tree_crown_detection}{GitHub} \\
\bottomrule
\end{tabular*}
\end{table*}

\subsection{Expert annotation and review}
\label{sec:sup_annotation_review}
Eight experts with interdisciplinary backgrounds in agriculture and
computer science annotated the data by scene family.
All had prior experience in image annotation and dataset development
and received the same training on the annotation workflow and platform.
They wrote each expression directly from the image without a template
and identified its target boxes from cues such as appearance, relative
position, and relationships among objects.
The authors reviewed each expression and its boxes against the original
image. When the wording or box selection was disputed, the authors
decided whether to retain, revise, or remove the record.

For queries with targets, the image had to support each referring cue,
and the target set had to contain exactly the instances that satisfied
the full description.
For target-absent queries, the authors inspected the entire image to
confirm that no instance met all the stated conditions, even when the
source annotations contained no matching box.

\subsection{Partitioning and record consistency}
We assigned images to splits in groups, keeping each image and all
its referring expressions together. We held out complete groups
from the training pool for validation. Our PhenoBench test subset
came from the native validation split, and our Brazil Urban Tree
test subset came from the published fold-0 test split.
The training split contains 6,353 query records over 4,724 images
(3,023 groups), the validation split contains 721 over 459 images
(110 groups), and the test split contains 3,007 over 2,221 images
(1,192 groups).

Group membership followed source metadata and links identified
during visual review. We kept images from the same CropAndWeed
recording session together, as well as crops from the same original
frame in PhenoBench and ACFR. For GWHD~2021, we grouped images by
acquisition domain and linked domains from the same site and season.
For OAM-TCD, we kept tiles from the same orthomosaic together.
For Dumortier, we grouped tiles with overlapping geographic
footprints. LaboroTomato groups also included frames captured by
the same camera within five seconds of one another. Images with
no recorded link to another image formed single-image groups.

We verified split consistency at both the record and image levels.
Specifically, no query ID, image ID, or group ID was shared across
splits, and both images in each recorded association were assigned
to the same split. We further checked for duplicate image files
by comparing SHA-256 hashes across splits and found no matches.

\subsection{Distribution of target cardinality}
Table~\ref{tab:sup_cardinality} shows the distribution of
ground-truth target counts across all 10,081 queries in AgroVG+.
We count an instance once for each query that refers to it.
This yields 12,261 target occurrences, with 7,318 from
single-target queries and 4,943 from multi-target queries.

Of the 1,870 multi-target queries, 1,567 (83.8\%) refer to
exactly two instances. The 40 queries with ten or more targets
account for 18.3\% of all target occurrences in multi-target
queries.

\section{Training and Decoding Details}
\label{sec:sup_impl}

\subsection{Optimization and trainable parameters}

We initialized ReSAG from Qwen3-VL-8B-Instruct~\cite{bai_qwen3vl_2025}
and applied LoRA~\cite{huLORALOWRANKADAPTATION2022} to the attention
and feed-forward projections of the language decoder. The vision
encoder, visual mergers, and pretrained language-model weights,
including the original input embeddings and output head, remained
frozen. We optimized the language LoRA adapters, the regional
projections $W_q$ and $W_v$, and the input embedding and output
vector of each of the two added action tokens. For \BOX{}, we
initialized these vectors from the corresponding pretrained rows
for \texttt{[}. For \END{}, we used the rows for the end-of-response
token. Each action is represented by a single token. Coordinate
strings use the original vocabulary and may span multiple tokens.

During teacher-forced training, the decoder predicted each answer
token from the image--expression prefix and the preceding
ground-truth answer tokens. For each query, we computed
$\mathcal{L}_{\mathrm{seq}}$ as the mean loss over answer tokens,
including actions, coordinates, and punctuation. We then combined
the two objectives as
$\mathcal{L}_{\mathrm{seq}}+\lambda_{\mathrm{reg}}\mathcal{L}_{\mathrm{reg}}$
for each query and averaged over the queries in each optimizer
update. Table~\ref{tab:sup_config} summarizes the configuration.

\begin{table*}[!t]
\captionsetup{singlelinecheck=false,skip=3pt}
\noindent\begin{minipage}[t]{\columnwidth}
\vspace{0pt}
\centering
\small
\caption{\textbf{Target cardinality in AgroVG+.} $K$ denotes the number of ground-truth boxes for a query. Counts cover all three splits.}
\label{tab:sup_cardinality}
\renewcommand{\arraystretch}{1.10}
\begin{tabular*}{\linewidth}{@{\extracolsep{\fill}}lrr@{}}
\toprule
\textbf{Target count $K$} & \textbf{Query records} & \textbf{Target occurrences} \\
\midrule
0 & 893 & 0 \\
1 & 7,318 & 7,318 \\
2 & 1,567 & 3,134 \\
3 & 192 & 576 \\
4 & 47 & 188 \\
5--9 & 24 & 142 \\
$\geq 10$ & 40 & 903 \\
\midrule
Total & 10,081 & 12,261 \\
\bottomrule
\end{tabular*}
\end{minipage}\hfill
\begin{minipage}[t]{\columnwidth}
\vspace{0pt}
\centering
\small
\caption{\textbf{ReSAG training and inference configuration.}}
\label{tab:sup_config}
\vspace{\baselineskip}
\footnotesize
\renewcommand{\arraystretch}{0.95}
\begin{tabularx}{\linewidth}{@{}l@{\hspace{5pt}}L@{}}
\toprule
\textbf{Setting} & \textbf{Value} \\
\midrule
LoRA rank $r$ / $\alpha$ & 32 / 64 \\
LoRA dropout & 0.05 \\
Regional projection size & $p=128$ \\
Temperature & $\tau=0.1$ \\
Regional loss weight & $\lambda_{\mathrm{reg}}=0.1$ \\
Optimizer & AdamW \\
Learning rate & $5\times10^{-5}$ \\
Weight decay & 0.01 \\
Schedule / warmup & Cosine decay / 144 updates \\
Training length & 6 epochs \\
Effective batch size & 8 queries \\
Precision / hardware & BF16 / two NVIDIA A100 GPUs \\
Maximum pixel budget & 524,288 \\
Decoding & Greedy, with key/value cache \\
Maximum new tokens & 4,096 \\
Training coordinate precision & Three decimal places \\
\bottomrule
\end{tabularx}
\end{minipage}
\end{table*}

\noindent\textbf{Checkpoint selection.}
We evaluated the checkpoint from each of the six epochs on all
721 validation queries. We compared the results in epoch order,
starting with epoch~1. A later checkpoint was eligible to replace
the current choice only if it matched or exceeded that checkpoint
on each of the four ESA scores for single- and multi-target queries
at IoU 0.50 and 0.75. For an eligible checkpoint, we preferred the
higher unweighted mean of the four scores. Ties were resolved by
higher target-absence accuracy (N-acc), then by earlier epoch.

\par\noindent\begin{minipage}{\linewidth}
\noindent\textbf{Input prompt.}
Training and inference used the same user message, with the image
followed by the text below. We supplied no separate system message.
The placeholder \texttt{\{query\}} denotes the English referring
expression.

\begingroup
\smallskip
\setlength{\leftskip}{0.6em}
\setlength{\rightskip}{0.6em plus 1fil}
\noindent\small\ttfamily
Locate every individual target referred to by the expression in the
image. For each matching target output <BOX> followed by
[x1,y1,x2,y2], using top-left and bottom-right coordinates normalized
to 0-1000 relative to the full image width and height. List boxes in
lexicographic coordinate order, one separate box per target without
duplicates. Finish with <END>. If no target matches, output only
<END>. Do not include labels, explanations, JSON wrapper, or
Markdown.\\
Expression: \{query\}
\par\endgroup
\end{minipage}\par\medskip

\subsection{Box serialization and constrained decoding}

For an image of width $W$ and height $H$, we normalize each pixel
coordinate $(x,y)$ to $(1000x/W,\,1000y/H)$. To construct the
training sequence, we round the normalized coordinates to three
decimal places and sort the boxes lexicographically by
$(x_1,y_1,x_2,y_2)$, where $(x_1,y_1)$ and $(x_2,y_2)$ denote
the top-left and bottom-right corners. The inference prompt requests
the same order. Evaluation uses one-to-one matching, so the order
of the predicted boxes does not affect the score.

Let $c(b_i)$ denote the tokenized coordinate string
\texttt{[x1,y1,x2,y2]} for the $i$th box in sorted order.
For $K\geq1$ targets, the supervised response is
\begin{equation}
    Y = \BOX\,c(b_1)\;\cdots\;\BOX\,c(b_K)\,\END .
\end{equation}
\begin{samepage}
For an empty target set, the response consists of the single
token \END{}.\par
\end{samepage}

At inference, we use greedy decoding with a key/value cache and
a limit of 4,096 new tokens per response. Decoding starts at an
action boundary. At each step, the decoder selects the
highest-scoring token allowed by its current state
(Table~\ref{tab:sup_decode}). The same vocabulary constraints apply
during teacher-forced training. The preceding ground-truth tokens
determine the state during training, and the generated tokens
determine it at inference.

\par\addvspace{\medskipamount}
\noindent\begin{minipage}{\linewidth}
\captionsetup{type=table}
\centering
\small
\caption{\textbf{BOX/END decoding states and transitions.}}
\label{tab:sup_decode}
\normalsize
\renewcommand{\arraystretch}{1.10}
\begin{tabularx}{\linewidth}{@{}lL@{}}
\toprule
\textbf{State} & \textbf{Allowed output and transition} \\
\midrule
Action boundary
    & Emit \BOX{} to start a coordinate body or \END{} to
      terminate the response. \\
\addlinespace[3pt]
Coordinate body
    & Generate from the original vocabulary $\mathcal{V}_0$ with
      both action tokens masked. Return to the action boundary
      when the decoded body ends with \texttt{]}. \\
\bottomrule
\end{tabularx}
\end{minipage}\par\addvspace{\medskipamount}

After generation, we parse the complete response and validate
every box. If the response fails to parse, contains an invalid
box, or reaches the token limit without emitting \END{}, the
entire query is scored as incorrect (Sec.~\ref{sec:sup_eval}).

\section{Evaluation Protocol}
\label{sec:sup_eval}

\subsection{One-to-one matching and exact-set accuracy}

Let $\widehat{\mathcal{B}}$ denote the predicted boxes and
$\mathcal{B}$ the ground-truth boxes associated with a query.
At IoU threshold $\eta$, we form a bipartite graph with an edge
between $\hat b \in \widehat{\mathcal{B}}$ and $b \in \mathcal{B}$
whenever $\operatorname{IoU}(\hat b,b) \geq \eta$.
Let $M_\eta$ be the cardinality of a maximum matching in this
graph. The exact-set indicator is
\begin{equation}
  C_\eta =
  \ind{\text{valid output}}\,
  \ind{M_\eta = |\widehat{\mathcal{B}}| = |\mathcal{B}|}.
  \label{eq:sup_exactset}
\end{equation}
An exact match covers every predicted and ground-truth box,
regardless of their order. Duplicate predictions enter the graph
as separate entries, so each ground-truth box can match at most
one of them. For a target-absent query, a valid empty prediction
gives $C_\eta=1$.

We report ESA@0.50 and ESA@0.75 by averaging $C_\eta$ separately
over single-target and multi-target queries. N-acc is the percentage
of target-absent queries with a valid empty prediction.

\newpage
\begin{samepage}
Overall,
denoted $\mathrm{Pr}@(F_1=1)$, gives every test query equal weight
at IoU 0.50. For the AgroVG+ test split,
\begin{equation}
  \mathrm{Overall} =
  \frac{2166 A_{\mathrm{s}} + 591 A_{\mathrm{m}}
        + 250 A_{\mathrm{n}}}{3007},
  \label{eq:sup_overall}
\end{equation}
where $A_{\mathrm{s}}$ and $A_{\mathrm{m}}$ are the single-target
and multi-target ESA@0.50 scores, respectively, and
$A_{\mathrm{n}}$ is N-acc. All scores are expressed as percentages.
Percentage-point differences are computed before rounding.\par
\end{samepage}

\subsection{Output parsing and validation}

For each model, we parse its output and convert the predicted boxes
to corner coordinates in the original image before computing IoU.
The matching evaluator uses these boxes as supplied, including
duplicates, and applies no further clipping or coordinate repair.
A missing prediction, a truncated generation, or a parsing failure
receives zero credit for the entire query. Such queries remain in
the denominator, including when the ground-truth set is empty.

For ReSAG, a valid response ends with \END{} within the token
budget. Each \BOX{} must be followed by exactly four finite
coordinates satisfying $0 \leq x_1 < x_2 \leq 1000$ and
$0 \leq y_1 < y_2 \leq 1000$. For an image of width $W$ and
height $H$, we scale horizontal coordinates by $W/1000$ and
vertical coordinates by $H/1000$. An initial \END{} gives a valid
empty prediction. A malformed coordinate tuple or a missing
terminator invalidates the entire response.

Table~\ref{tab:sup_baseline_parsing} gives the output format and
empty-prediction rule used for each baseline. The two detectors'
score thresholds were selected on the validation split and applied
before IoU matching. For zero-shot Ferret, we rescored the saved
responses with its native coordinate parser after observing that
the JSON parser failed on its free-form outputs.

\section{Additional Analyses and Visualizations}
\label{sec:sup_analysis}

\subsection{Accuracy by scene family and target count}

\noindent\textbf{Scene family.}
At IoU 0.50, ReSAG achieves higher exact-set accuracy than the
Qwen3-VL-8B LoRA baseline in five of the six scene families
(Table~\ref{tab:sup_scene_results}). The largest gain is
4.62 percentage points on wheat heads. ReSAG answers 337 of
379 pest queries correctly, three fewer than LoRA. Tree canopies
have the lowest accuracy for both models.

\begingroup
\looseness=-1
\noindent\textbf{Target count.}
ReSAG improves on LoRA in every target-count group
(Table~\ref{tab:sup_target_count_results}). For both models,
exact-set accuracy declines as the number of targets increases.
ReSAG recovers the complete set for 76.03\% of two-target queries
and 58.90\% of three-target queries. Among the 34 queries with
four or more targets, ReSAG recovers 11 complete sets and
LoRA recovers 10.\par
\endgroup

\onecolumn
\noindent\begin{minipage}{\textwidth}
\captionsetup{type=table}
\centering
\small
\caption{\textbf{Baseline output parsing on AgroVG+.}
All boxes are mapped to the original image for scoring.
The last column specifies when a model output counts as a valid
empty prediction.}
\label{tab:sup_baseline_parsing}
\footnotesize
\setlength{\tabcolsep}{4pt}
\begin{tabularx}{\textwidth}{
  @{}>{\raggedright\arraybackslash}p{0.22\textwidth}
  L
  >{\raggedright\arraybackslash}p{0.225\textwidth}@{}}
\toprule
\textbf{Model} &
\textbf{Box extraction and coordinate conversion} &
\textbf{Valid empty prediction} \\
\midrule

MDETR-GREC
& Retain detections with $1-p(\text{no object}) \geq 0.90$.
  Convert centers and sizes to corners and clip to the image bounds.
& No detection meets the score threshold. \\

Grounding DINO-T
& Retain detections whose maximum sigmoid score over text positions
  exceeds $0.40$. Convert centers and sizes to corners and clip to
  the image bounds.
& No detection exceeds the score threshold. \\

Florence-2-large-ft
& Decode groups of four location tokens with the native
  phrase-grounding processor using the original image dimensions.
& Generation terminates with an empty decoded body. \\

Shikra-7B (LoRA)
& Read bracketed boxes in $[0,1]$ relative to the square-padded
  image. Remove the padding offsets and clip the boxes to the
  original image bounds.
& An explicit \texttt{[]}. \\

Qwen3-VL-8B (zero-shot)
& Read the JSON \texttt{boxes} field on the $[0,1000]$ scale.
  A flat list of four numbers in this field is also accepted as
  one box. Scale by the original image dimensions.
& An empty list in the \texttt{boxes} field. \\

Qwen3-VL-8B, InternVL3.5-8B, and Ferret-7B (LoRA)
& Read the list of boxes from the JSON \texttt{boxes} field.
  Scale the $[0,1000]$ corner coordinates by the original image
  dimensions.
& An empty list in the \texttt{boxes} field. \\

Ferret-7B (zero-shot)
& Extract all bracketed groups of four integer coordinates in
  $[0,1000]$. Map them to integer pixel coordinates with the
  native decoder.
& A complete, nonempty response containing no square brackets. \\

\bottomrule
\end{tabularx}
\end{minipage}\par\medskip

\noindent\begin{minipage}[t]{\dimexpr(\textwidth-\columnsep)/2\relax}
\vspace{0pt}
\captionsetup{type=table}
\centering
\small
\caption{\textbf{Exact-set accuracy by scene family.}
Scores are percentages at IoU 0.50 over all test queries in each
family, including target-absent queries. Bold marks the higher score.}
\label{tab:sup_scene_results}
\footnotesize
\begin{tabularx}{\linewidth}{@{}Lrrr@{}}
\toprule
\textbf{Scene family} & \textbf{Queries} & \textbf{LoRA} & \textbf{ReSAG} \\
\midrule
Crops and weeds & 541 & 81.52 & \textbf{84.66} \\
Fruits          & 553 & 81.56 & \textbf{83.00} \\
Wheat heads     & 390 & 79.23 & \textbf{83.85} \\
Pests           & 379 & \textbf{89.71} & 88.92 \\
Plant diseases  & 468 & 73.93 & \textbf{74.79} \\
Tree canopies   & 676 & 55.47 & \textbf{57.25} \\
\midrule
All queries     & 3,007 & 75.22 & \textbf{77.09} \\
\bottomrule
\end{tabularx}
\end{minipage}\hfill
\begin{minipage}[t]{\dimexpr(\textwidth-\columnsep)/2\relax}
\vspace{0pt}
\captionsetup{type=table}
\centering
\small
\caption{\textbf{Exact-set accuracy by ground-truth target count.}
Scores are percentages at IoU 0.50. The zero-target row reports
N-acc. Bold marks the higher score.}
\label{tab:sup_target_count_results}
\footnotesize
\begin{tabularx}{\linewidth}{@{}Lrrr@{}}
\toprule
\textbf{Targets} & \textbf{Queries} & \textbf{LoRA} & \textbf{ReSAG} \\
\midrule
0        &   250 & 82.80 & \textbf{83.60} \\
1        & 2,166 & 76.45 & \textbf{77.89} \\
2        &   484 & 72.31 & \textbf{76.03} \\
3        &    73 & 53.42 & \textbf{58.90} \\
$\geq 4$ &    34 & 29.41 & \textbf{32.35} \\
\bottomrule
\end{tabularx}
\end{minipage}
\par\smallskip

\subsection{Qualitative grounding examples}

Of the two reddish strawberries in
Figure~\ref{fig:sup_single_target_examples}, ReSAG selects the one
to the right of the lower green berry and below and to the left of
the large red berry. For the group requested in
Figure~\ref{fig:sup_multi_target_examples}, ReSAG returns all three
apples clustered on the left and excludes the isolated apple on the
right. Both figures show predicted coordinates normalized to
$[0,1000]$ along each image axis and rounded to integers for display.
\par\smallskip\noindent\begin{minipage}{\textwidth}
\captionsetup{type=figure,skip=5pt}
\centering
  \includegraphics[width=\textwidth]{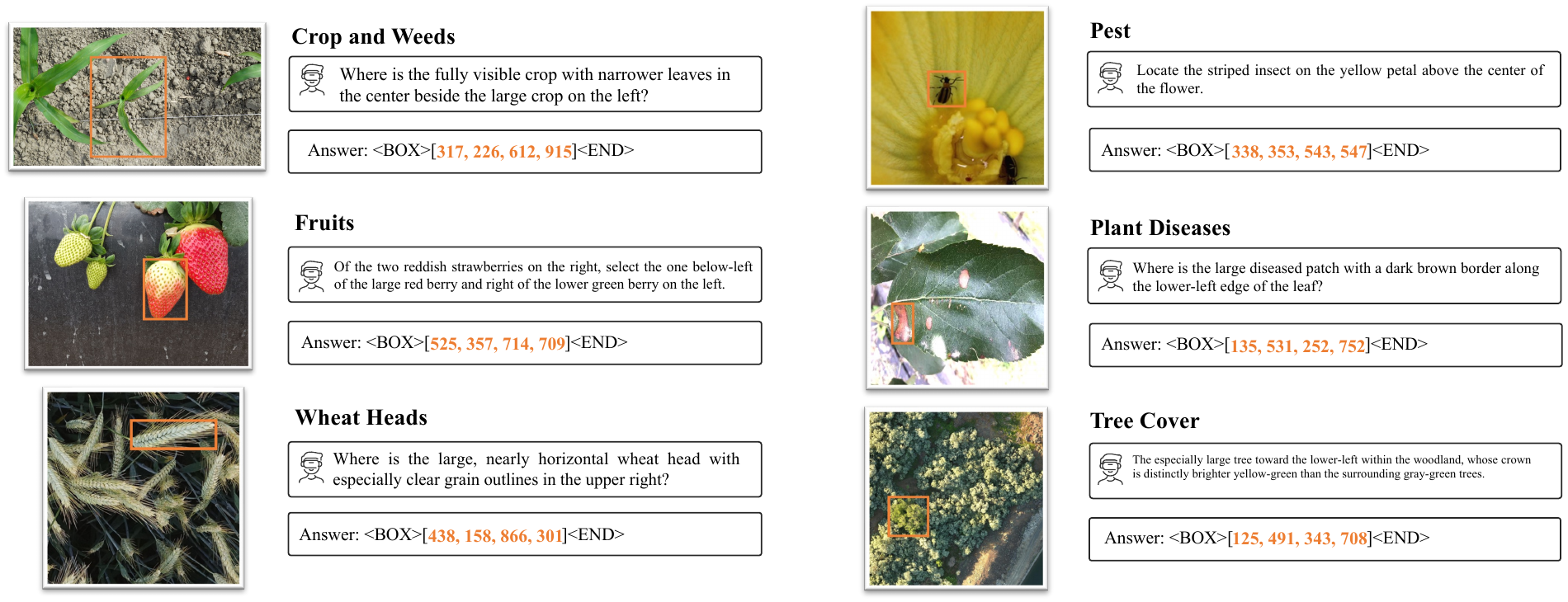}
  \caption{\textbf{Single-target grounding across six agricultural
  scene families.} Each orange box marks the instance selected by
  ReSAG for the accompanying English test query. The answer below
  the query gives the predicted corner coordinates between the
  \BOX{} and \END{} actions.}
  \label{fig:sup_single_target_examples}
\end{minipage}

\twocolumn[{
\begin{minipage}{\textwidth}
\captionsetup{type=figure,skip=5pt}
\centering
  \includegraphics[width=\textwidth]{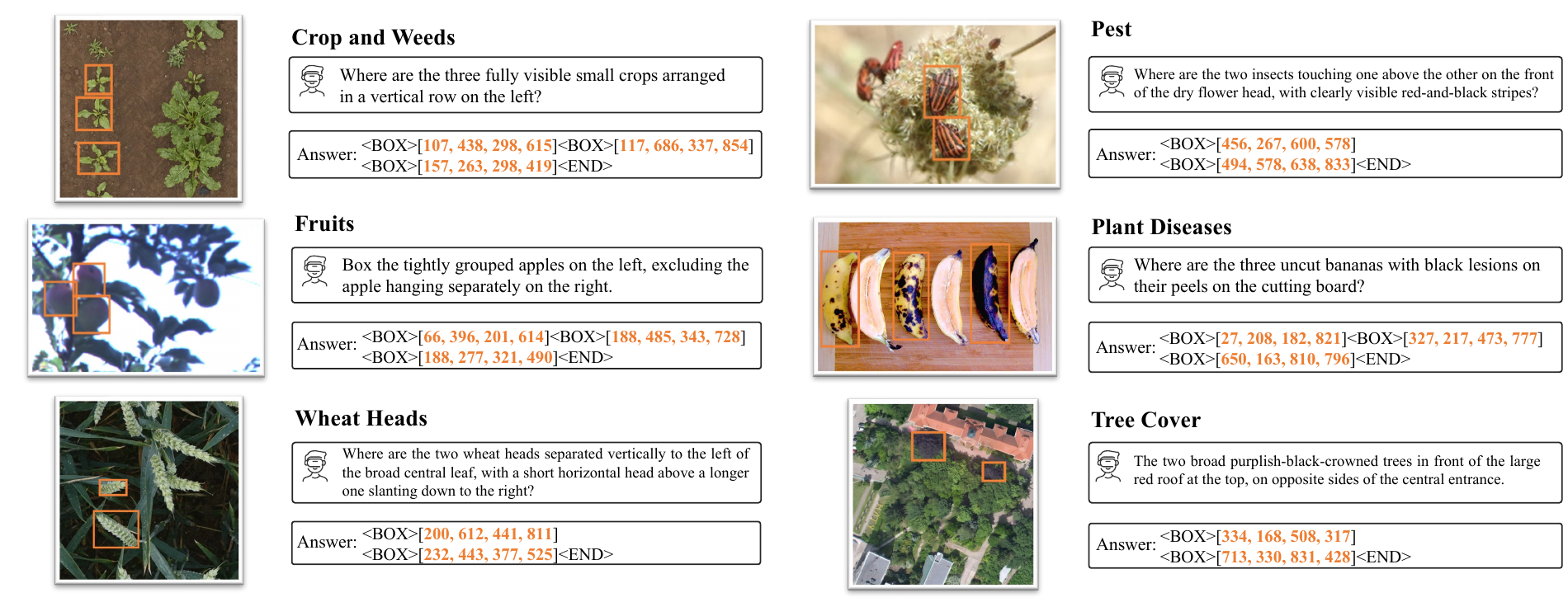}
  \caption{\textbf{Multi-target grounding across the same six scene
  families.} Orange boxes show the predicted set. Each \BOX{} action
  introduces one box, and \END{} terminates the response.}
  \label{fig:sup_multi_target_examples}
\par\smallskip
\end{minipage}
}]

\subsection{Failure analysis}
\label{sec:sup_failures}

In 557 of its 689 errors at IoU 0.50 (80.8\%), ReSAG predicts
the correct number of boxes, but the boxes cannot be matched
one-to-one to the full target set. LoRA makes 611 errors of this
type. On target-present queries, ReSAG predicts too few boxes
in 77 cases and too many in 14. The remaining 41 errors are
nonempty predictions on target-absent queries. All 3,007 ReSAG
responses pass the output checks in Sec.~\ref{sec:sup_eval}.

\begin{figure}[!htbp]
  \centering
  \includegraphics[width=0.85\linewidth]{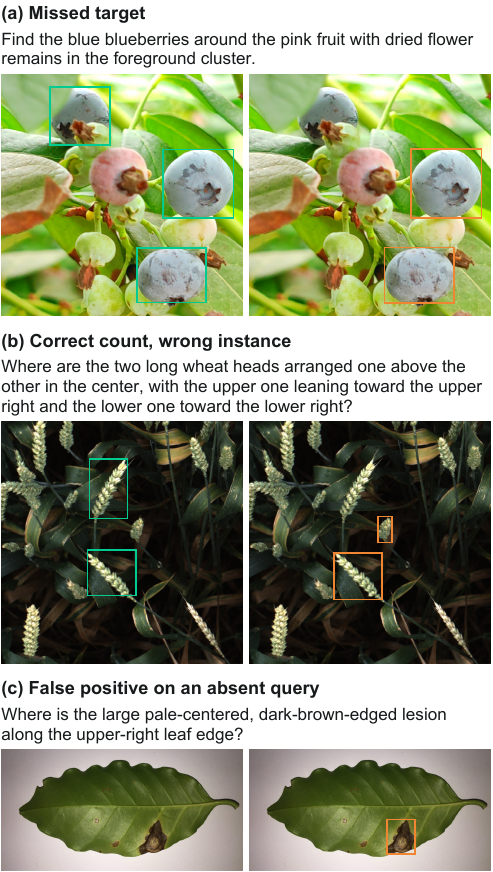}
  \caption{\textbf{Three failures of complete-set grounding.}
  Each pair shows the ground truth (left, green) and ReSAG predictions
  (right, orange) for the query above it.
  The crop in (a) retains every annotated and predicted box.}
  \label{fig:sup_failure_cases}
\end{figure}

Figure~\ref{fig:sup_failure_cases} shows three of these errors.
ReSAG misses the upper-left blueberry in (a). In (b), it locates
the lower of the two requested wheat heads but places its second
box on a shorter head to their right. In (c), the query specifies
a lesion along the upper-right leaf edge, while the predicted box
encloses a dark patch on the lower edge.

\noindent\textbf{Scope of interpretation.}
The regional loss pools evidence within each annotated box, which
can include background or parts of neighboring instances. Our
evaluation covers English queries on the AgroVG+ test split.
Grounding in Chinese and transfer to new farms, sensors, or seasons
require separate evaluation.

\section*{Code and Data Availability}
The ReSAG code is available on \AgroVGReleaseLink{GitHub}{} and
\AgroVGReleaseLink{Hugging Face}{}. The accompanying AgroVG+ dataset
includes bilingual referring expressions, query-specific target boxes,
and training, validation, and test splits.

\end{document}